\documentclass[lettersize,journal]{IEEEtran}
\usepackage{amsmath,amsfonts}
\usepackage{algorithmic}
\usepackage{algorithm}
\usepackage{array}
\usepackage{bbm}
\usepackage{booktabs}
\usepackage[caption=false,font=normalsize,labelfont=sf,textfont=sf]{subfig}
\usepackage{textcomp}
\usepackage{stfloats}
\usepackage{url}
\usepackage{verbatim}
\usepackage{graphicx}
\usepackage{cite}
\usepackage{cleveref} 
\crefname{figure}{figure}{figure}
\crefname{table}{Table}{Table}
\usepackage{multirow}
\usepackage{enumitem}
\begin{document}

\title{Non-Destructive Detection of Sub-Micron Imperceptible Scratches On Laser Chips Based On Consistent Texture Entropy Recursive Optimization Semi-Supervised Network}

\author{Pan Liu
\thanks{Pan Liu is with the School of Automation, Central South University, Changsha 410083, China (e-mail: pan.liu@csu.edu.cn)}}

\maketitle

\begin{abstract}
Laser chips, the core components of semiconductor lasers, are extensively utilized in various industries, showing great potential for future application. Smoothness emitting surfaces are crucial in chip production, as even imperceptible scratches can significantly degrade performance and lifespan, thus impeding production efficiency and yield. Therefore, non-destructively detecting these imperceptible scratches on the emitting surfaces is essential for enhancing yield and reducing costs. These sub-micron level scratches, barely visible against the background, are extremely difficult to detect with conventional methods, compounded by a lack of labeled datasets. To address this challenge, this paper introduces TexRecNet, a consistent texture entropy recursive optimization semi-supervised network. The network, based on a recursive optimization architecture, iteratively improves the detection accuracy of imperceptible scratch edges, using outputs from previous cycles to inform subsequent inputs and guide the network's positional encoding. It also introduces image texture entropy, utilizing a substantial amount of unlabeled data to expand the training set while maintaining training signal reliability. Ultimately, by analyzing the inconsistency of the network output sequences obtained during the recursive process, a semi-supervised training strategy with recursive consistency constraints is proposed, using outputs from the recursive process for non-destructive signal augmentation and consistently optimizes the loss function for efficient end-to-end training. Experimental results show that this method, utilizing a substantial amount of unsupervised data, achieves 75.6$\%$ accuracy and 74.8$\%$ recall in detecting imperceptible scratches, an 8.5$\%$ and 33.6$\%$ improvement over conventional Unet, enhancing quality control in laser chips.
\end{abstract}


\section{Introduction}
\begin{figure}[!t]
\centering
\includegraphics[width=1.0\columnwidth]{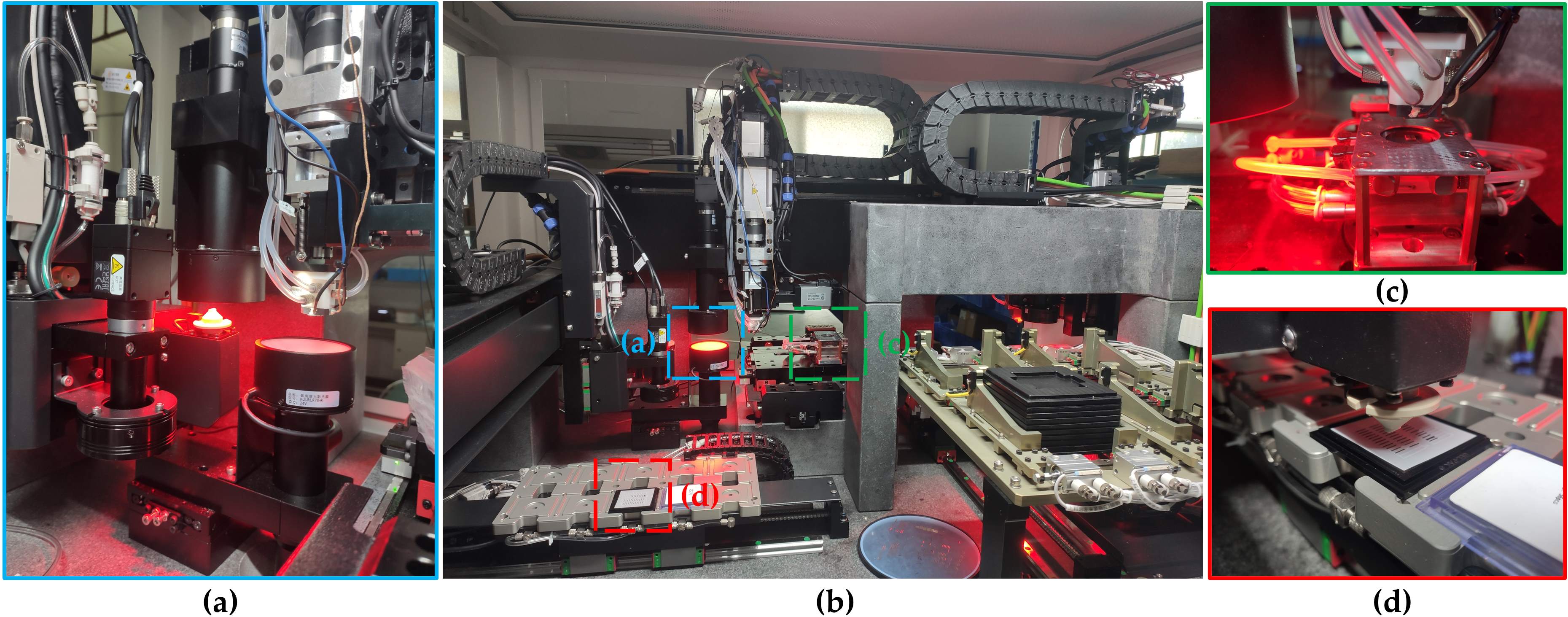}
\caption{Laser Chip Manufacturing Site (a) Mechanical Calibration (b) Production scenario (c) Eutectic Bonding (d) Chip Retrieval}
\label{fig_1}
\end{figure}

\IEEEPARstart{L}{aser} chips, optoelectronic devices essential for laser emission, are crucial in both civilian and military applications including medical diagnostics, consumer electronics, and LIDAR systems\cite{ref1}. As depicted in \cref{fig_1}, laser chip production involves complex procedures such as substrate preparation, edge growth, etching, cleaving, and die bonding. Frequent interactions between mechanical arms and chips during this process commonly causes surface damage such as scratches\cite{ref2}. Scratches on non-emitting areas of the chips can lead to problems like light leakage and scattering, disrupting illumination in complex laser devices and potentially leading to equipment failure. Therefore, accurately locating and identifying scratch defects on chips is crucial for diagnosing chip malfunctions, reducing production costs, and ensuring high yields in production\cite{ref3}.
\begin{figure}[!t]
\centering
\includegraphics[width=1.0\columnwidth]{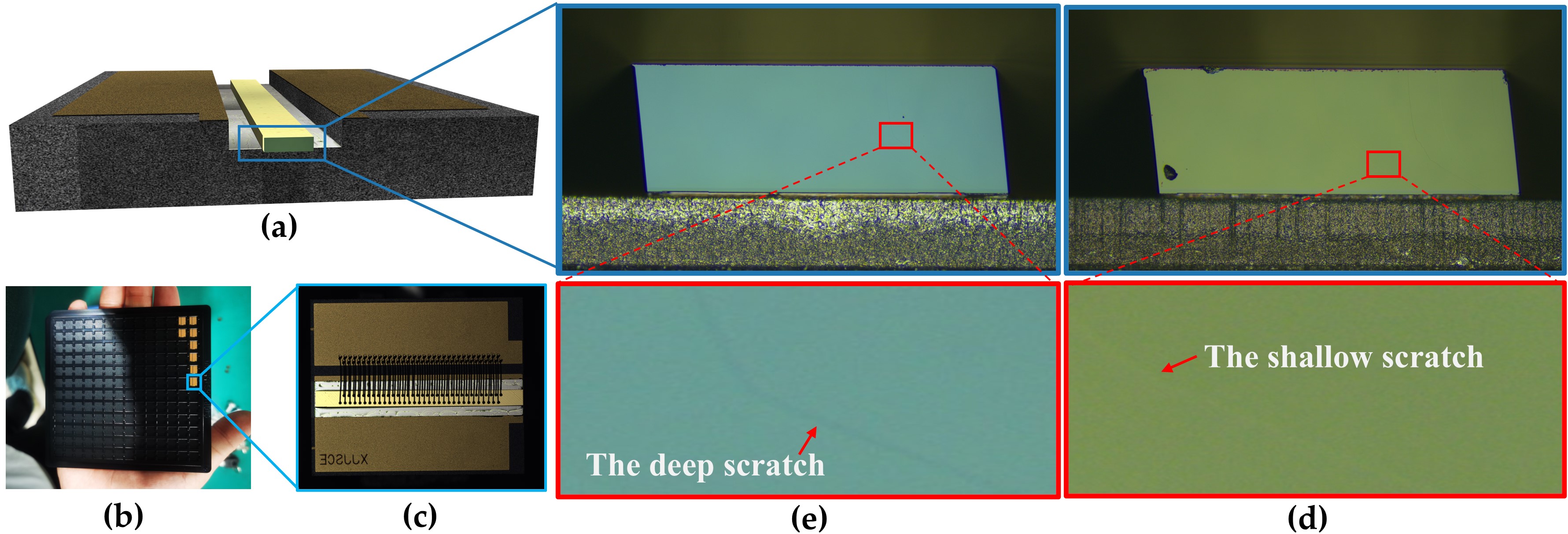}
\caption{Schematic diagram of laser chip image and location of scratches (a) 3D structure of chip, (b) Chip general view, (c) Chip enlargement view, (d) deep scratches schematic diagram, (e) shallow scratches schematic diagram}
\label{fig_2}
\end{figure}
Detecting surface scratches on laser chips is a significant challenge in the field of micro-nano optoelectronics\cite{ref4}. The key lies in employing high-precision image analysis techniques, like semantic segmentation, to distinguish and categorize image pixels for identifying defects\cite{ref5}. However, unlike traditional image segmentation tasks such as identifying scratches on metal surfaces\cite{ref6} or lesion in medical imaging\cite{ref7}, detecting chip scratches presents two distinct challenges. As shown in \cref{fig_2}, firstly, scratch defects typically exist only on a sub-micrometer scale\cite{ref8}. Their optical similarity to the surrounding substrate makes them hard to distinguish, particularly shallow scratches that account for over 20$\%$ of these defects and have extremely low contrast, making them almost invisible. Consequently, the minuscule size and low contrast of these scratch defects greatly diminishes the accuracy of automated segmentation algorithms. Secondly, the lack of distinct identifying features in scratch defects makes creating a large, accurately labeled dataset challenging\cite{ref9}, crucial for training efficient detection models with supervised learning\cite{ref10}. Therefore, while semantic segmentation is promising, further development of advanced recognition models and effective training strategies is essential for precise and reliable automatic scratch detection in laser chips\cite{ref11}.

In response to challenges posed by low-contrast images, extensive research has focused on supervised semantic segmentation models. Challenges include unclear target boundaries, low signal-to-noise ratios, and faint features. Traditional image segmentation techniques like threshold-based methods are somewhat effective but less so for low-contrast and complex background images\cite{ref12}. Modern deep learning techniques, especially Convolutional Neural Networks (CNNs), have significantly enhanced segmentation model performance in these conditions\cite{ref13}. Mu et al. addressed this by extracting multiple low-level features and obtaining saliency information for low-contrast images through inter-covariance computation, serving as prior data to achieve low-contrast segmentation \cite{ref14}. Tabernik D et al. employed a Unet\cite{ref15} with attention mechanisms for detecting surface defect under low-contrast conditions, yielding significant results\cite{ref16}. Zhou et al. introduced a multi-scale densely connected high-resolution encoder-decoder network, acquiring high-resolution semantic information crucial for accurate boundary localization\cite{ref17}. Hou et al. further improved sensitivity to subtle image features and segmentation accuracy by introducing a complex self-attention mechanism design\cite{ref18}. Heidari et al. utilized Transformer\cite{ref19} and CNN for multi-scale feature extraction from low-contrast images, achieving a refined integration of global and local features\cite{ref20}.

Semi-supervised semantic segmentation techniques are considered a potent approach to address the challenge of insufficient labeled datasets for laser chip scratches. This method involves training models with limited labeled data, supplemented by a larger set of unlabeled data to enhance model generalizability\cite{ref21}. These approaches improve segmentation accuracy and also greatly reduce the time and costs of manual labeling\cite{ref22}. Various semi-supervised semantic segmentation methods have been recently proposed. GANs\cite{ref23}, using pseudo-label generators and discriminators, assist in constructing supervisory signals for unlabeled images\cite{ref24}. U2PL, segregates reliable from unreliable parts in erroneous pseudo-labels for comprehensive data utilization, expanding pseudo-labeling\cite{ref25}. ST++ discards low-reliability pseudo-label masks entirely to reduce cognitive confusion from incorrect labels\cite{ref26}, analyzing pseudo-label reliability at the image level. CPS has devised a cross pseudo-training framework with cross-entropy losses between different network outputs, simple yet effective\cite{ref27}. The EMA teacher model-based consistency regularization method achieves semi-supervised training with consistency loss between teacher and student model outputs\cite{ref28}. CutMix proposes an image augmentation method that combines cutting and sample mixing to enhance consistency differences\cite{ref29}. FitMatch integrates regularization and self-training, using weakly perturbed images for pseudo-label generation to guide the training of strongly perturbed images\cite{ref30}. IMAS manages consistency training by varying perturbation levels based on images and predictions\cite{ref31}.

Despite advancements, current semi-supervised segmentation methods still face numerous challenges in identifying scratches on laser chips.

\textbf{Network structure non-specificity:} In semi-supervised tasks, ideal supervision signal construction depends on accurate initial predictions. However, traditional models often yield unreliable predictions due to the subtle scale of sub-micrometer scratch defects and their low contrast characteristics, which closely resemble the substrate. This issue is exacerbated under limited annotated data training, leading to cognitive confusion.
\textbf{Detachment of semi-supervised signal construction from the object features:} Current semi-supervised semantic segmentation methods excessively depend on the confidence space to construct supervisory signals. Once cognitive confusion occurs in the model, the interpretability of the confidence space significantly decreases, further degrading supervisory signal quality. 
\textbf{Sensitivity to the signal-to-noise ratio in image enhancement:} Image enhancement techniques commonly used in semi-supervised semantic segmentation might disrupt the original image structure. For scratch data, slight contrast differences between the foreground and background amplify this structural disruption, leading to loss of structural information and directional biases in supervisory signal construction.

\begin{figure*}[!h]
\centering
\includegraphics[width=2\columnwidth]{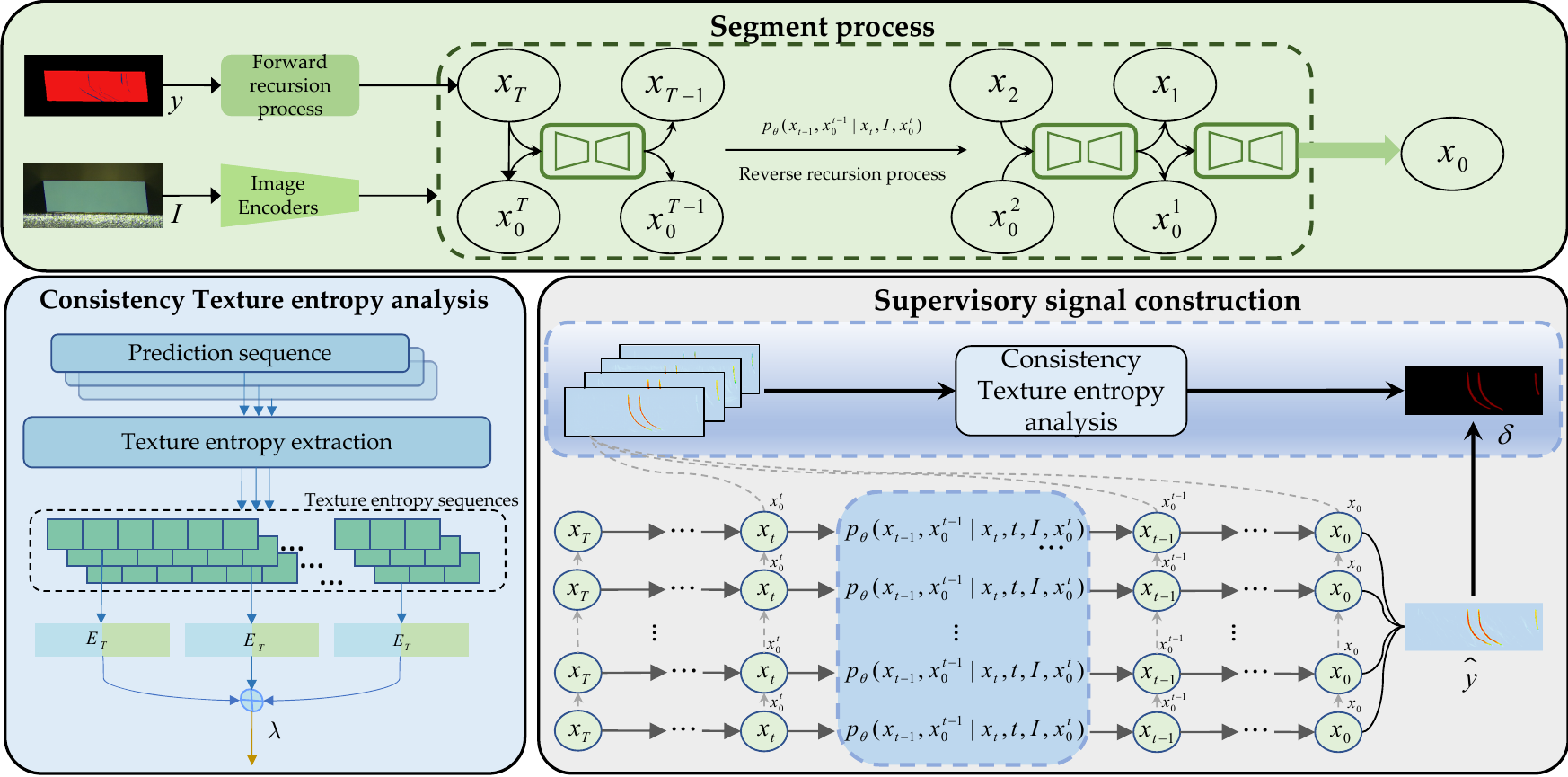}
\caption{Overview of TexRecNet pipeline for unlabeled images.The highly reliable prediction is obtained by the recursive optimized prediction process, and then the reliable supervision signal is extracted by texture entropy consistency.}
\label{fig_3}
\end{figure*}

To overcome the main challenges encountered in semi-supervised learning for detecting imperceptible scratch defects at the sub-micrometer level on laser chip surfaces, this study introduces TexRecNet, a consistent texture entropy recursive optimization semi-supervised network. This network is designed to address the instability of scratch segmentation and improve supervision signal construction, particularly under conditions of limited labeled data and extensive unlabeled data. The architecture of this study is detailed as follows:\textbf{1) }A recursive optimization network structure has been developed, optimizing segmentation predictions recursively. High-confidence segmentation masks generated from preceding recursions are used as prior information in subsequent recursions, helping the model iteratively refine defect structures and significantly enhance scratch detection accuracy, especially in low contrast conditions.\textbf{2) }A supervision signal construction method based on texture entropy has been proposed. This method involves analyzing the texture structure features of imperceptible scratches and designing an image texture entropy that describes edge texture distribution. This quantifies texture differences in recursive predictions of unlabeled data and combined with prediction sequence confidence, constructs a highly reliable supervision signal.\textbf{3) }A semi-supervised training strategy with a recursive consistency constraint has been developed. By constructing recursive prediction sequences through random initial sampling, it achieves lossless data augmentation of training signals. Based on the non-consistency features between prediction sequences, a consistency constraint loss function is constructed, enabling fast end-to-end training with unlabeled data.

The primary contributions of this paper include:

\begin{enumerate}[leftmargin=*,align=left]
\item{\textit{Developing a recursive optimization network architecture}: This involves using high-confidence predictions from previous recursions as priors to enhance model recognition capabilities}
\item{\textit{Introducing a texture entropy-based supervision signal construction method}: This quantitatively assesses the network's ability to recognize subtle features, facilitating reliable augmentation of the training dataset with unlabeled data.}
\item{\textit{Formulating a semi-supervised training strategy with a recursive consistency constraint}: This achieves end-to-end semi-supervised training through lossless augmentation and sequence consistency loss, minimizing the reliance on repetitive training.}
\end{enumerate}

The remainder of the paper is structured as follows: Chapter II details the proposed TexRecNet; Chapter III discusses experimental results and analyses; Chapter IV concludes the work and results of this paper.
\section{Methodology}

This section provides a detailed overview of the core mechanisms and structural configuration of TexRecNet. As illustrated in \cref{fig_3}, TexRecNet, a recursive optimization scratch segmentation network, is proposed to address the sub-micrometer scale of scratch defects and their extremely low contrast with the foreground and background. Based on the idea of recursive optimization, the network uses predictions from prior iterations as inputs for subsequent ones, thereby progressively refines segmentation predictions, easing feature extraction under low-contrast conditions. Following this refined approach, an efficient semi-supervised training strategy is further developed to mitigate the challenges posed by the scarcity of large-scale annotated scratch datasets. This strategy involves constructing sets of recursive prediction sequences for unlabeled data through random initial sampling, then integrating prediction confidence with the proposed texture entropy to generate reliable supervision signals. Thereby, end-to-end efficient training is achieved by employing a consistency-constrained loss function.
\subsection{TexRecNet}
\begin{figure*}[!t]
\centering
\includegraphics[width=2\columnwidth]{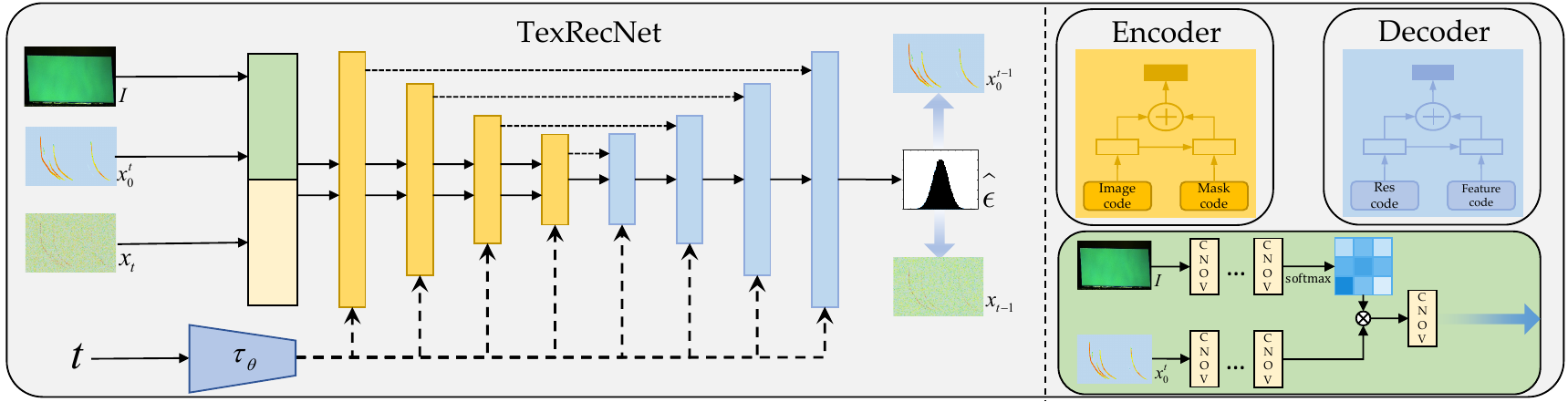}
\caption{TexRecNet Network Architecture. Images and historical predictions are input simultaneously for encoding and fused through attention structures.}
\label{fig_4}
\end{figure*}

Precise differentiation of foreground and background features in semantic segmentation is crucial for model performance. This is especially true in laser chip image processing, where there is significant overlap in representing scratch defects and chip backgrounds. Traditional methods for extracting low-contrast features depend exclusively on information from image inputs. However, their effectiveness diminishes significantly in scenarios with diminutive targets and low contrast in chip image. As show in \cref{fig_4}, TexRecNet, a network structure optimized recursively, is proposed to address the aforementioned challenges. The segmentation process is reconstructed as a finite recursive process. The network iteratively optimizes the initial noisy input masks, ultimately converting them into desired location predictions. Simultaneously, the process involves sampling previous prediction results across time steps to construct foreground-background position data, calibrating the subsequent feature localization of the network. This stage-by-stage refinement and location priorization enable the framework to surpass traditional feature extraction challenges in chip images, thus improving the accuracy of network predictions.

a) \textbf{Running Process:} Following the current multi-step encoder-decoder network paradigm\cite{ref32}, the operational process of our network is decomposed into forward and reverse recursive processes. During the forward recursive process $q$, Gaussian noise is progressively added to the label across time steps $T$, in accordance with a pre-specified variance table $\left\{ 0,{{\alpha }_{1}},{{\alpha }_{2}},...,{{\alpha }_{T}} \right\}$, generating a sequence of noisy masks $\left\{ {{x}_{0}},{{x}_{1}},...,{{x}_{T}} \right\}$,
\begin{equation}
\label{deqn_ex1}
q({{x}_{0:T}}|y):=\Pi _{t=1}^{T}\mathcal{N}({{x}_{t}};\sqrt{{{\alpha }_{t}}}{{x}_{t-1}},(1-{{\alpha }_{t}}){{I}_{n\times n}})
\end{equation}
Where ${{I}_{n\times n}}$ represents an identity matrix of size $n$. Furthermore, the forward process can be directly sampled at any time step $t$,
\begin{equation}
\label{deqn_ex2}
\setlength\abovedisplayskip{6pt}
\setlength\belowdisplayskip{6pt}
{{x}_{t}}=\sqrt{{{\overline{\alpha }}_{t}}}y\!+\!\sqrt{1-{{\overline{\alpha }}_{t}}}\epsilon ,\epsilon \sim \mathcal{N}(0,{{I}_{n\times n}}),s.t.{{\overline{a}}_{t}}=\prod\limits_{s}^{t}{{{a}_{s}}}.
\end{equation}

In the reverse recursive process ${{p}_{\theta }}$, parameterized by $\theta$, the objective is to incrementally remove noise from ${{x}_{T}}$, retracing the recursive process in reverse $\left\{ {{x}_{T}},{{x}_{T-1}},...,{{x}_{0}} \right\}$ to obtain the final prediction ${{x}_{0}}$. To construct a prior signal, it is assumed that the network ${{\varepsilon }_{\theta }}$ has gained essential foreground-background localization information ${{\widehat{x}}_{\text{0}}}=x_{0}^{t}$ at any time step $t$. Utilizing the current mask ${{x}_{t}}$ and $x_{0}^{t}$ to be optimized, ${{\varepsilon }_{\theta }}$ is responsible for predicting noise component $\widehat{\epsilon }={{\varepsilon }_{\theta }}({{x}_{t}},t,I,x_{0}^{t})$ in ${{x}_{t}}$ and reducing the noise in ${{x}_{t}}$, based on $\widehat{\epsilon }$, to obtain ${{x}_{t-1}}$. This iterative process is repeated until the final refined output is obtained
\begin{equation}
\label{deqn_ex3}
\begin{split}
{{x}_{t-1}}&=\alpha _{t}^{-\frac{1}{2}}\left( {{x}_{t}}-\frac{1-{{\alpha }_{t}}}{\sqrt{1-{{\overline{\alpha }}_{t}}}}{{\varepsilon }_{\theta }}({{x}_{t}},t,I,x_{0}^{t}) \right)\\
           &+\!{{\mathbbm{1}}_{[t>1]}}(1-{{\alpha }_{t}}){{I}_{n\times n}}
\end{split}.
\end{equation}

Simultaneously, the basic localization $x_{0}^{t}$ is updated based on the following equation to gradually optimize the prior signal, enhancing the network's predictive outcomes
\begin{equation}
\label{deqn_ex4}
x_{0}^{t-1}\!:=\! 
\begin{cases}
&\hspace{-1em}\overline{\alpha }_{t}^{-\frac{1}{2}}\left( {{x}_{t}}\!-\!\sqrt{1-{{\overline{\alpha }}_{t}}}{{\varepsilon }_{\theta }}({{x}_{t}},t,I,x_{0}^{t+1}) \right),t\!<\!T \\ 
&\hspace{-1em} 0,t=T\\ 
\end{cases}\!.
\end{equation}

Consequently, the entire reverse recursive process ${{p}_{\theta }}$ can be expressed as
\begin{equation}
\label{deqn_ex5}
\hspace{-0.5em}{{p}_{\theta }}\!({{x}_{0:T\!-\!1}},\!x_{0}^{0:T\!-\!1}\!|{{x}_{T}},\!T,\!I,\!x_{0}^{T})\!\!:=\!\!\Pi _{t\!=\!1}^{T}{{p}_{\theta }}({{x}_{t\!-\!1}}\!,\!x_{0}^{t\!-\!1}\!|{{x}_{t}},\!t,\!I,\!x_{0}^{t}).\!
\end{equation}

Thereupon, the inference process can be defined as shown in \cref{alg:alg1}.
\begin{algorithm}[H]
\caption{Inference Algorithm.}\label{alg:alg1}
\begin{algorithmic}[1]
\STATE ${{x}_{T}}\sim \mathcal{N}(0,{{I}_{n\times n}}),x_{0}^{T}=0$
\FOR{$t=T,...,1$} 
\STATE $\widehat{\epsilon }={{\varepsilon }_{\theta }}({{x}_{t}},t,I,x_{0}^{t})$
\STATE $x_{0}^{t-1}=\overline{\alpha }_{t}^{-\frac{1}{2}}\left( {{x}_{t}}-\sqrt{1-{{\overline{\alpha }}_{t}}}\widehat{\varepsilon } \right)$
\STATE ${{x}_{t-1}}=\alpha _{t}^{-\frac{1}{2}}\left( {{x}_{t}}-\frac{1-{{\alpha }_{t}}}{\sqrt{1-{{\overline{\alpha }}_{t}}}}\widehat{\varepsilon } \right)+{{\mathbbm{1}}_{[t>1]}}(1-{{\alpha }_{t}}){{I}_{n\times n}}$
\ENDFOR
\RETURN ${{x}_{0}}$
\end{algorithmic}
\label{alg1}
\end{algorithm}

\textbf{b) Network Structure:} At the time step $t$, the network input comprises the mask to be optimized ${{x}_{t}}\in {{\mathbb{R}}^{H\times W\times 1}}$, the original chip image $I\in {{\mathbb{R}}^{H\times W\times C}}$, and the prior mask ${{\widehat{x}}_{\text{0}}}\in {{\mathbb{R}}^{H\times W\times 1}}$.

In this setup, both the image $I$ and the prior mask ${{\widehat{x}}_{\text{0}}}$ serve as conditional inputs, undergoing convolution through multiple 2D layers to obtain C-channel feature maps. For improved guidance of the encoding process through mask prediction, the feature map of the prior mask ${{\widehat{x}}_{\text{0}}}$ is subjected to specialized layer normalization. This normalized map is then multiplied with the feature map obtained from $I$ to construct conditional feature maps, focusing areas of interest. These conditional feature maps are then processed by an encoder $R$ with multiple residual layers, resulting in improved encoding. Meanwhile, the main input ${{x}_{t}}$ retains its original scale after being encoded through multiple 2D convolutional layers. This main input, together with the conditional feature maps, is then input into the encoder-decoder $U$ through residual connections.

$U$ is constructed as a cross-linked encoder-decoder architecture. Notably, the time embedding acquired at the current time step $t$ through the encoder ${{\tau }_{\theta }}$ is incorporated as a learnable embedding into $U$. In contrast, the network structure that encodes the input and constructs the attention as part of the precoding, does not incorporate time embeddings. This design renders these components time-insensitive, enabling them to adapt to predictions of ${{\widehat{x}}_{\text{0}}}$ from any input.
\subsection{Network Training Process}
For laser chip scratch detection, a small labeled dataset ${{D}_{l}}=\left\{ {{I}_{i}},{{y}_{i}} \right\}_{i=1}^{\left| {{D}_{l}} \right|}$ and a significantly larger unlabeled dataset ${{D}_{u}}=\{{{I}_{u}}\}_{u=1}^{\left| {{D}_{u}} \right|}$ are available. The objective is to train the network utilizing the extensive amount of unlabeled data, supplemented by a limited amount of labeled data. The crucial aspect of training the unlabeled set ${{D}_{u}}$ is the construction of supervisory signals. Due to the cost constraints of manual labeling, using only manually labeled tags as universal signals is impractical. Therefore, a method for constructing supervisory signals based on texture entropy has been proposed.

\textbf{a) Training with Labeled Data:} During each training iteration, a random image ${{I}_{i}}$ and its corresponding label ${{y}_{i}}$ are extracted from the labeled dataset ${{D}_{l}}$, and a time step $t$ within $\left[ 0,T \right]$ is randomly assigned. ${{y}_{i}}$ is sampled as ${{x}_{t}}$ using \cref{deqn_ex2}, and the training loss is constructed as
\begin{equation}
\label{deqn_ex6}
{{E}_{{{x}_{0}},\epsilon ,t}}[\|\epsilon -{{\varepsilon }_{\theta }}(\sqrt{{{\overline{a}}_{t}}}{{x}_{0}}+\sqrt{1-{{\overline{a}}_{t}}}\epsilon ,t,{{I}_{i}},{{\widehat{x}}_{0}}){{\|}^{2}}].
\end{equation}

Since ${{\widehat{x}}_{0}}$ cannot be directly obtained without the network, the supervised training of TexRecNet is decomposed into a coupled training process. Initially, ${{\widehat{x}}_{0}}$ is set as a non-biased mask $0$ for a basic prediction $x_{0}^{t-1}$ , and then, ${{\widehat{x}}_{0}}$ is assumed to be $x_{0}^{t-1}$ for a subsequent prediction. Finally, the network trains on the losses from both predictions simultaneously. Despite the inconsistency of ${{\widehat{x}}_{0}}$ in time steps during training and inference, the network maintains strong generalizability due to its time-insensitive conditional encoding. This procedure is detailed in the supervised section of \cref{alg:alg2} Furthermore, the final supervised loss ${{\mathcal{L}}_{l}}$ is defined as follows,
\begin{equation}
\label{deqn_ex7}
\begin{cases}
& \hspace{-1em}{{\mathcal{L}}_{l}}\!=\!{{E}_{{{x}_{0}},\epsilon ,t}}[\|\epsilon\! -\!{{\widehat{\epsilon }}_{1}}|{{|}^{2}}\!+\!\|\epsilon \!-\!{{\widehat{\epsilon }}_{2}}{{\|}^{2}}] \\ 
& \hspace{-1em}{{\widehat{\epsilon }}_{1}}\!=\!{{\varepsilon }_{\theta }}(\sqrt{{{\overline{a}}_{t}}}{{x}_{0}}\!+\!\sqrt{1\!-\!{{\overline{a}}_{t}}}\epsilon ,t,{{I}_{i}},0) \\ 
& \hspace{-1em}{{\widehat{\epsilon }}_{2}}\!=\!{{\varepsilon }_{\theta }}(\sqrt{{{\overline{a}}_{t}}}{{x}_{0}}\!+\!\sqrt{1\!-\!{{\overline{a}}_{t}}}\epsilon ,t,{{I}_{i}},\overline{\alpha }_{t}^{-\frac{1}{2}}\left( {{x}_{t}}\!-\!\sqrt{1\!-\!{{\overline{\alpha }}_{t}}}{{\widehat{\epsilon }}_{1}} \right)) \\ 
\end{cases}
\end{equation}

\textbf{b) Constructing Supervisory Signals for Unlabeled Data:} For mask predictions of the same instance ${{I}_{u}}$, variations in initial ${{x}_{T}}$ or the time steps $t$ lead to divergent results of $x_{0}^{t}$. This divergence is more noticeable in instances that demonstrate poor convergence. As a result, the method emphasizes extracting and quantitatively describing changes in variability across different prediction sequences to guide the construction of supervision signals.

To minimize time and storage requirements in constructing supervisory signals, the reverse inference sequence $\left\{ {{x}_{T}},{{x}_{T-1}},...,{{x}_{0}} \right\}$ is evenly divided into $M$ parts. From each part, a sample is randomly selected to construct a new inference sequence $\left\{ {{x}_{M}},{{x}_{M-1}},...,{{x}_{0}} \right\}$, significantly shortening the inference process. Then, $N$ instances of ${{x}_{M}}$ are sampled from $\mathcal{N}(0,{{I}_{n\times n}})$ to create a sequence $\{x_{M}^{1},x_{M}^{2},...,x_{M}^{N}\}$, enhancing the prediction diversity while maintaining the original image distribution. Thus, a rich yet concise prediction sequence is obtained,
\begin{equation}
\label{deqn_ex8}
{{\left\{ x_{0,i}^{1},...,x_{0,i}^{M} \right\}}_{\!1:N\!}}\!\gets\!{{p}_{\theta }}{{({{x}_{0:M\!-\!1}}\!,x_{0}^{0:M\!-\!1}|x_{M}^{i}\!,M\!,I\!,x_{0}^{M})}_{\!1:N}}.\!
\end{equation}
$\{x_{0,1}^{1},x_{0,2}^{1},...,x_{0,N}^{1}\}$ is generally deemed more reliable than other predictions, making it the primary component of the supervisory signal $\widehat{y}$
\begin{equation}
\label{deqn_ex9}
\widehat{y}=\mathbbm{1}\!\| \overline{x}_{0}^{1}\ge {{\tau }_{m}}\|,s.t.\overline{x}_{0}^{j}=\frac{1}{N}\sum\limits_{i=1}^{N}{x_{0,i}^{j}}.
\end{equation}
Where ${{\tau }_{m}}$ is the segmentation threshold. However, using only the final prediction set to construct supervisory signals is too simplistic. To address this, based on the prediction sequence ${{\{x_{0,i}^{1},,...,x_{0,i}^{M}\}}_{1:N}}$, a loss mask $\delta$ is constructed to filter out ineffective pixels,
\begin{equation}
\label{deqn_ex10}
\delta\!=\!\mathbbm{1}\!\|\overline{x}_{0}^{1}\!\ge\! (1\!-\!\lambda ){{\tau }_{c}}\|\mathbbm{1}\!\|{{E}_{j\!=\!1}^{M}}({{E}_{i\!=\!1}^{N}}\|x_{0,i}^{j}\!-\!\overline{x}_{0}^{j}\|)\!\!\ge\!\! (1\!-\!\lambda ){{\tau }_{v}}\|.
\end{equation}
Where ${{\tau }_{c}}$ and ${{\tau }_{v}}$ are the confidence and consistency thresholds, respectively. Their purpose is to restrict the generation of supervisory signals and loss masks, delineating high-fidelity representations to construct highly precise loss masks.$\lambda $ symbolizes the consistent texture entropy feature, designed to regulate the generation of loss masks from the feature space.

Thus, the unsupervised loss ${{\mathcal{L}}_{u}}$ can be formulated,
\begin{equation}
\label{deqn_ex11}
\setlength\abovedisplayskip{3pt}
\setlength\belowdisplayskip{3pt}
\hspace{-0.6em}{{\mathcal{L}}_{u}}\!=\!\sum\limits_{j}^{N}{\delta }\|{{(1\!-\!{{\overline{\alpha }}_{t}})}^{\!-\!\frac{1}{2}}}\sum\limits_{i}^{M}{{({{x}_{i,j}}\!-\!\overline{\alpha }_{t}^{\frac{1}{2}}\widehat{y}\!-\!{{\widehat{\epsilon }}_{i,j}})}/{(M\!\times\! N)}\;}\!\!{{\|}^{2}}\!.
\end{equation}

Additionally, the approach for consistent texture entropy feature extraction is detailed below. To extract differences in edge features across prediction sequences, a quantitative description of the predicted foreground structure is essential. Therefore, based on the significant divergences in the subtle destruction of mask texture structures across different predictions, the concept of texture entropy ${{E}_{T}}$ is introduced. This concept aims to capture subtle structural changes in predictions, providing a quantitative description of the predicted mask distribution representation.

Firstly, to extract the structural texture features, prediction ${{x}_{0}}$ is transformed into a structural feature space
\begin{equation}
\label{deqn_ex12}
\vartheta ({{x}_{0}})\!:=\!B\otimes {{x}_{0}},s.t.B\!=\!\left[ \begin{matrix}
   {{k}^{0}} & ... & {{k}^{{{K}_{s}}}}  \\
   ... & ... & ...  \\
   {{k}^{({{K}_{s}}\!-\!1)\times {{K}_{s}}}} & ... & {{k}^{{{K}_{s}}\!\times\! {{K}_{s}}}}  \\
\end{matrix} \right].
\end{equation}
Where $\otimes$ represents convolution,$B$ is the information extraction matrix,$k$ is the number of segmentation categories, and ${{k}_{s}}$ is the size of the structural perception domain.

Then, the structural encoding $S$ is defined as
\begin{equation}
\label{deqn_ex13}
S({{x}_{0}}):=\frac{1}{\left\| \vartheta ({{x}_{0}}) \right\|}\!
\begin{bmatrix}
 \sum{\left\| \mathbbm{1}\!\left[ \vartheta ({{x}_{0}})=0 \right] \right\|}\\
...\\
\sum{\left\| \mathbbm{1}\!\left[ \vartheta ({{x}_{0}})={{k}^{{{K}_{s}}\times ({{K}_{s}}+1)}} \right] \right\|}\\
\end{bmatrix}\!.\!
\end{equation}
Based on the level of disarray in the structural encoding, texture entropy is defined as follows,
\begin{equation}
\label{deqn_ex14}
{{E}_{T}}({{x}_{0}})=-S{{({{x}_{0}})}^{T}}\times {{\log }_{2}}S({{x}_{0}}).
\end{equation}

Based on the aforementioned texture entropy ${{E}_{T}}$, we can quantitatively describe the complexity of texture structures within prediction results. Theoretically, during the prediction process, a gradual optimization of the input time step $t$ and the prior mask $x_{0}^{t}$ is expected to lead to a progressive decrease in the complexity of the prediction results. However, in actual prediction scenarios, this reduction in complexity is discontinuous. This discontinuity stems from the network's inherent cognitive bias regarding the input image. Therefore, identifying high-frequency shifts in texture entropy within the prediction mask sequences quantifies this discontinuity, reflecting the calibration level of the network to the input image and measuring the actual reliability of its predictions. As a result, consistent texture entropy features are defined as follows,
\begin{equation}
\label{deqn_ex15}
\lambda \!:=\!E\!\!\left( \frac{\sum\nolimits_{f\!>\!{{\tau }_{f}}}\!{\left| \mathcal{F}\!\{{{E}_{T}}(\!{{x}_{0}}\!),{{E}_{T}}(\!{{x}_{1}}\!),...,{{E}_{T}}(\!{{x}_{M}}\!)\}(f) \right|}}{\sum\nolimits_{f}{\left| \mathcal{F}\!\{{{E}_{T}}(\!{{x}_{0}}\!),{{E}_{T}}(\!{{x}_{1}}\!),...,{{E}_{T}}(\!{{x}_{M}}\!)\}(f) \right|}} \right)\!.\!
\end{equation}
Where $\mathcal{F}$ denotes the Fourier transform operation,${{\tau }_{f}}$ is the high-frequency component threshold, and $\mathcal{F}\text{ }\!\!\{\!\!\text{ }\cdot \text{ }\!\!\}\!\!\text{ (}f\text{)}$ represents the value at frequency $f$ after the Fourier transform of the object.

\textbf{c) Training Framework:} As illustrated in \cref{alg:alg2}, our multi-step segmentation framework, which deviates from conventional semi-supervised semantic segmentation approaches, generates extra prediction mask sequences $\left\{ {{x}_{0}},x_{0}^{1},x_{0}^{2},...,x_{0}^{T-1} \right\}$ for unlabeled data ${{I}_{u}}$ during the inference process. This approach enriches the information available for semi-supervised training process. Consequently, a training method based on recursive prediction sequences is adopted. Recursive prediction sequence sets are derived through resampling, and consistency constraint losses among these sets are constructed. This integration of unlabeled data supervision signal construction and network inference aims to avoid model retraining, thus significantly reducing training overhead. Furthermore, it prevents the loss incrued from disrupting the original image distribution.

\section{Experiment}
In this section, the effectiveness of the proposed method on a laser chip scratch dataset will be validated. Additionally, extensive ablation studies will be conducted to examine the contributions of each module to model performance enhancement.

\textbf{Dataset.} Experiments were conducted using a dataset of chip scratch defects. Optical and other equipment were installed in a laser chip co-bonding patch scenario at a chip processing factory to capture images of chip defects. The dataset was comprised of 363 non-repetitive images, each measuring $3200 \times 1700$ pixels, showcasing scratch defects. Images were cropped to $2560 \times 1024$ to capture main chip areas, and subsequently augmented to 3630 images of $256 \times 256$ through $1/2$ scaling, cropping, and other enhancement techniques. Of these images, 2240 were used as an unlabeled dataset to evaluate the generalization ability of the model. The remaining 1390 were carefully annotated pixel-by-pixel as background, shallow, or deep scratch defects using professional labeling tools, forming the labeled dataset. The labeled dataset was further divided into 1082 training, 154 validation, and 154 test images, aiming for comprehensive training and evaluation of the model.

\begin{algorithm}[!t]
\caption{Training Algorithm.}\label{alg:alg2}
\begin{algorithmic}[1]
\REPEAT 
\STATE \textbf{Sample}$({{I}_{l}},{{y}_{l}})\sim{{D}_{l}}$ // unlabeled set training
\STATE $\epsilon \sim\mathcal{N}(0,{{I}_{n\times n}}),t\sim\{0,...,T\}$
\STATE ${{x}_{t}}=\sqrt{{{\overline{\alpha }}_{t}}}(2y-1)+\sqrt{1-{{\overline{\alpha }}_{t}}}\epsilon $
\STATE ${{\widehat{\epsilon }}_{1}}={{\varepsilon }_{\theta }}({{x}_{t}},t,I,0)$
\STATE $x_{0}^{t-1}=\overline{\alpha }_{t}^{-\frac{1}{2}}\left( {{x}_{t}}-\sqrt{1-{{\overline{\alpha }}_{t}}}{{\widehat{\epsilon }}_{1}} \right)$
\STATE ${{\widehat{\epsilon }}_{2}}={{\varepsilon }_{\theta }}({{x}_{t}},t,I,x_{0}^{t-1})$
\STATE $\nabla \left[ {{\left\| \epsilon -{{\widehat{\epsilon }}_{1}} \right\|}^{2}}+{{\left\| \epsilon -{{\widehat{\epsilon }}_{2}} \right\|}^{2}} \right]$
\STATE \textbf{Sample}${{I}_{u}}\sim{{D}_{u}}$ // unlabeled set training
\FOR{$i=1,..,N$}
\STATE $\epsilon \sim\mathcal{N}(0,{{I}_{n\times n}}),t\sim\{\frac{M-1}{M}T,...,T\}$
\STATE ${{x}_{M,i}}=\sqrt{{{\overline{\alpha }}_{t}}}(2y-1)+\sqrt{1-{{\overline{\alpha }}_{t}}}\epsilon $
\STATE $\left\{ \{x_{0,i}^{1},...,x_{0,i}^{M}\},\{{{\widehat{\epsilon }}_{1,i}},...,{{\widehat{\epsilon }}_{M,i}}\},\{{{x}_{1,i}},...,{{x}_{M,i}}\} \right\}$\\
$\gets{{p}_{\theta }}({{x}_{0:M-1}},x_{0,i}^{0:M-1}|{{x}_{M,i}},M,I,x_{0,i}^{M})$
\ENDFOR
\STATE $(\widehat{y},\delta )=Eval({{\left\{ x_{0,i}^{1},...,x_{0,i}^{M} \right\}}_{1:N}})$
\STATE $\nabla \sum\limits_{j}^{N}{\delta }\|{{(1\!-\!{{\overline{\alpha }}_{t}})}^{-\frac{1}{2}}}\sum\limits_{i}^{M}{{({{x}_{i,j}}\!-\!\overline{\alpha }_{t}^{\frac{1}{2}}\widehat{y}\!-\!{{\widehat{\epsilon }}_{i,j}})}/{(M\times N)}\;}\!{{\|}^{2}}$
\UNTIL{convergence}
\end{algorithmic}
\label{alg2}
\end{algorithm}

\textbf{Implementation Details.} In this study, the model was initialized with random parameters as weights. The sampling step length $T$ was set to 100 to reduce inference time. All experiments utilized the AdamW optimizer for end-to-end  network training. The initial learning rate was set to $1\times {{10}^{-4}}$. Input images were uniformly enhanced with a 50$\%$ probability of random flipping, random scaling within $\left[ 0.8,1.2 \right]$, rotation within$\left[ {{0}^{o}},{{360}^{o}} \right]$, and normalization. During unsupervised training, we set the sampling sequence length $M$ to 12, and the number of sampling sequences $N$ to 2, enhancing sampling diversity.${{\tau }_{m}}$ was set to 0.5 to facilitate pseudo-label generation. To obtain high-quality loss masks, we set${{\tau }_{c}}=0.8,{{\tau }_{v}}=0.05$ and${{\tau }_{f}}=9$. Moreover, due to the significant proportion of background compared to shallow and deep scratches in the dataset, shallow and deep scratch classes were unified into a single scratch class during actual training, and scratch loss was amplified fivefold to mitigate the issue of insufficient foreground data. Additionally, experiments utilized a sliding window for segmentation to obtain continuous image results.

\textbf{Evaluation Metrics.} The chip scratch defect dataset exhibits a significant class imbalance, with background pixels outnumbering foreground pixels by nearly a hundredfold. Given that predicting background is easier than predicting foreground, the effectiveness of background prediction varies little across different methods. Therefore, foreground Intersection over Union (IoU), Dice coefficient, and accuracy are used as universal metrics for evaluating prediction results. Additionally, shallow scratch recall rate and deep scratch recall rate are employed as supplementary metrics to evaluate the capability of the network in detecting shallow and deep scratch defects.
\subsection{Comparison with existing alternative methods}
\begin{figure*}[!t]
\centering
\includegraphics[width=2\columnwidth]{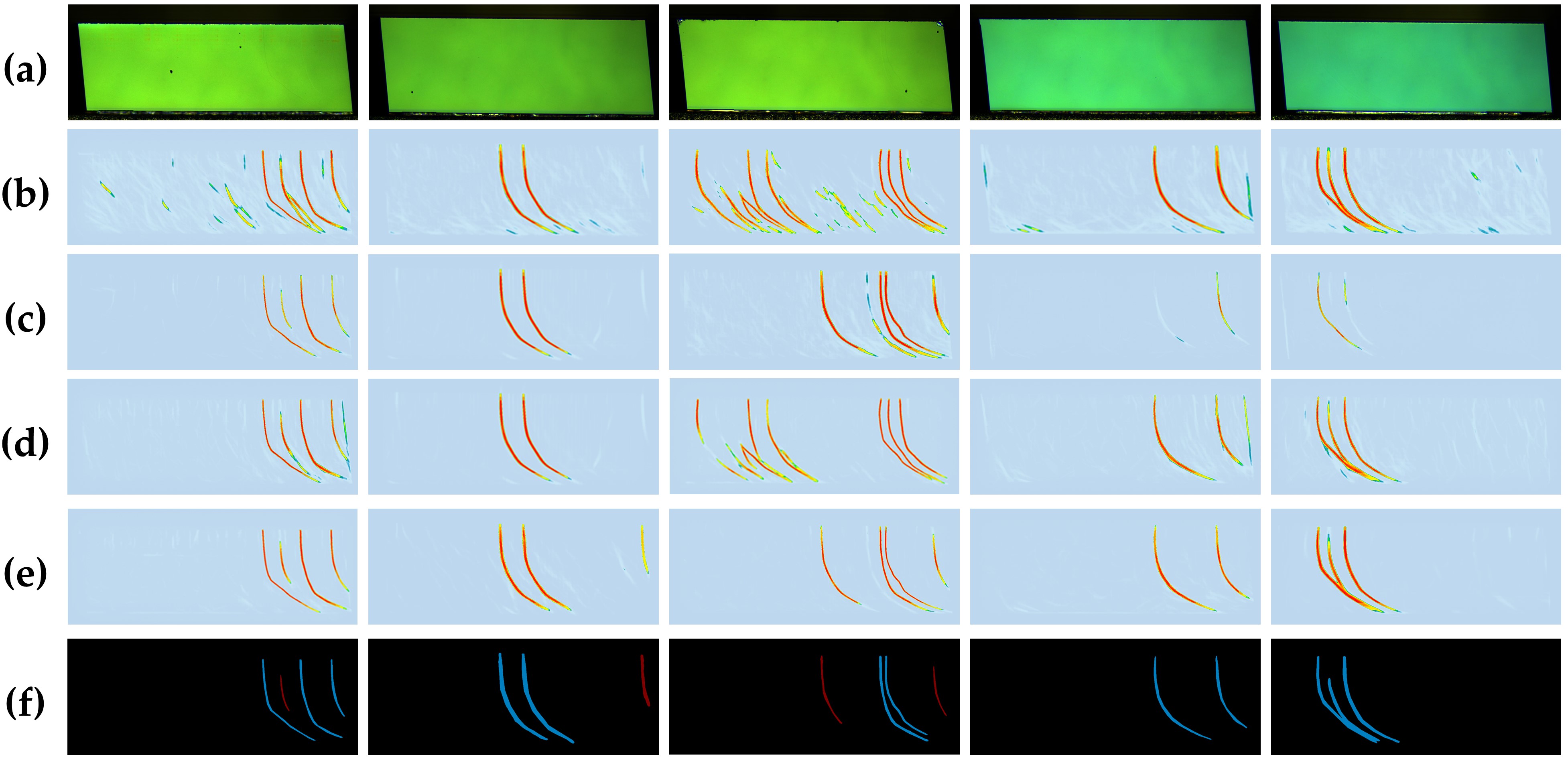}
\caption{Performance demonstration of different methods (a) Chip collection image (b) Unet prediction result (c) CutMix prediction result (d) MT prediction result (e) Prediction result of our method (f) Ground truth}
\label{fig_5}
\end{figure*}

\textbf{Overall Accuracy Comparison.} To empirically analyze the effectiveness of the proposed method, Unet\cite{ref15}, Segmenter\cite{ref33}, DANet\cite{ref34}, and SegDeff\cite{ref35}, were used as validated supervised semantic segmentation models for comparative analysis. Furthermore, MT\cite{ref28}, CutMix\cite{ref29}, and iMAS\cite{ref31} were chosen as benchmarks to further evaluate the performance of the method in semi-supervised semantic segmentation. Specifically, to investigate the impact of different semi-supervised data partitions on model performance, a subset of data is randomly selected from the labeled dataset, delabeled, and added to the unlabeled dataset. For a balanced comparison, we present a performance analysis of TexRecNet against other advanced methods in segmentation for chip scratch defect segmentation. Test results are illustrated in \cref{fig_5} and \cref{tab1}, with shallow and deep scratches marked in red and blue, respectively.

Firstly, the method in this paper, under supervised conditions, outperformed the traditional fully supervised Unet approach, exhibiting a 4.6$\%$ increase in foreground IoU. Although the results are very similar to MedSegDiff in deep scratch recall rate, which employs multi-step optimization, the proposed model still shows a 2.5$\%$ improvement in IoU and 3.2$\%$ in the Dice coefficient. Moreover, it significantly surpasses this benchmark with a 17.2$\%$ higher shallow scratch recall rate. Comparative experiments demonstrate that traditional deep networks in a fully supervised context struggle with scratch location representation, mainly due to feature blurring between foreground and background in chip images, a problem especially prominent in detecting shallow scratches. Our solution, TexRecNet, diverges from other deep networks in two key aspects: (a) It breaks down the segmentation into a finite serious of recursive predictions, each optimized iteratively by the network to enhance detection accuracy progressively. (b) During the iteration process, it extracts position data of the foreground and background from previous predictions, using this data as a guide for aiding encoder capturing essential information. The experimental results thoroughly substantiate the logic and effectiveness of these design features.

\begin{table}
\centering
\caption{Comparison of statistical errors among various models.}
\label{tab1}
    \begin{tabular}{ccccccc}
    \toprule
    \toprule
    \multirow{2}[4]{*}{Method}
    & \multicolumn{2}{c}{dataset partition} 
    & \multirow{2}[4]{*}{IoU} 
    & \multirow{2}[4]{*}{Dice} 
    & \multirow{2}[4]{*}{\shortstack{Shallow \\ Recall}} 
    & \multirow{2}[4]{*}{\shortstack{Deep \\ Recall}} \\
\cmidrule{2-3}          & Labeled & Unlabeled &       &       &       &       \\
    \midrule
    Unet      & 1082  & ―     & 67.1  & 80.3  & 41.2  & 70.5 \\
    Segmenter & 1082  & ―     & 66.8  & 80.0  & 42.8  & 71.3 \\
    DANet     & 1082  & ―     & 67.9  & 80.8  & 47.5  & 72.4 \\
    MedsegDiff& 1082  & ―     & 69.2  & 81.7  & 46.2  & 73.8 \\
    ours      & 1082  & ―     & \textbf{71.7} & \textbf{83.5} & \textbf{63.4} & \textbf{76.4} \\
    \midrule
    MT     & 1082  & 2240(1:2) & 71.1  & 83.1  & 65.8  & 79.3 \\
    CutMix & 1082  & 2240(1:2) & 73.2  & 84.5  & 64.9  & 80.9 \\
    iMAS   & 1082  & 2240(1:2) & 74.1  & 85.1  & 67.5  & 81.4 \\
    ours   & 1082  & 2240(1:2) & \textbf{75.6} & \textbf{86.1} & \textbf{74.8} & \textbf{82.6} \\
    MT     & 664   & 2658(1:4) & 67.6  & 80.6  & 54.2  & 75.9 \\
    CutMix & 664   & 2658(1:4) & 72.3  & 83.9  & 57.5  & 76.4 \\
    iMAS   & 664   & 2658(1:4) & 73.0  & 84.3  & 60.8  & 78.2 \\
    ours   & 664   & 2658(1:4) & \textbf{74.5} & \textbf{85.3} & \textbf{70.2} & \textbf{80.3} \\
    MT     & 369   & 2953(1:8) & 66.1  & 79.5  & 53.8  & 71.2 \\
    CutMix & 369   & 2953(1:8) & 70.6  & 82.7  & 52.7  & 73.5 \\
    iMAS   & 369   & 2953(1:8) & 70.2  & 82.4  & 57.6  & 74.6 \\
    ours   & 369   & 2953(1:8) & \textbf{72.4} & \textbf{83.9} & \textbf{68.3} & \textbf{77.1} \\
    \bottomrule
    \bottomrule
    \end{tabular}
\end{table}

Furthermore, the model presented in this paper served as the network architecture for comparing the effectiveness of various semi-supervised methods. As depicted in \cref{tab1}, the proposed method consistently outperformed both classic and advanced semi-supervised methods, showing significant performance advantages across different dataset partitions. It maintained a nearly 10$\%$ lead in shallow scratch recall rate and also excelled in other performance metrics. This improvement is attributed to the proposed consistency constraint semi-supervised training strategy, which constructs different recursive starting points through random sampling and, based on recursive optimization, generates prediction sequence sets. This strategy effectively augments supervision signals, all the while preserving the original low signal-to-noise ratio characteristics of shallow scratches. Additionally, despite the significantly reduction in labeled data, the proposed method preserved high detection accuracy, achieving impressive foreground IoU scores of 72.4$\%$, 74.5$\%$, and 75.6$\%$ across three different data partitions ($1/2$, $1/4$, $1/8$). Notably, with just 369 labeled images in a $1/8$ data partition, our method achieved a remarkable 14.5$\%$ increase in shallow scratch recall rate over the semi-supervised baseline MT, and a 10.7$\%$ increase over the latest iMAS. This result validates the low dependency of the model output on the proposed supervision signal construction method. This method utilizes texture entropy for constructing supervision signal, integrating texture differences and prediction confidence from various prediction sequences to collectively assess segmentation masks. This approach produces highly reliable unsupervised data supervision signals, resolving the issue of traditional methods that depend on single reference sources and often lead to erroneous biased supervision signals.

\begin{figure}[!t]
\centering
\includegraphics[width=1.0\columnwidth]{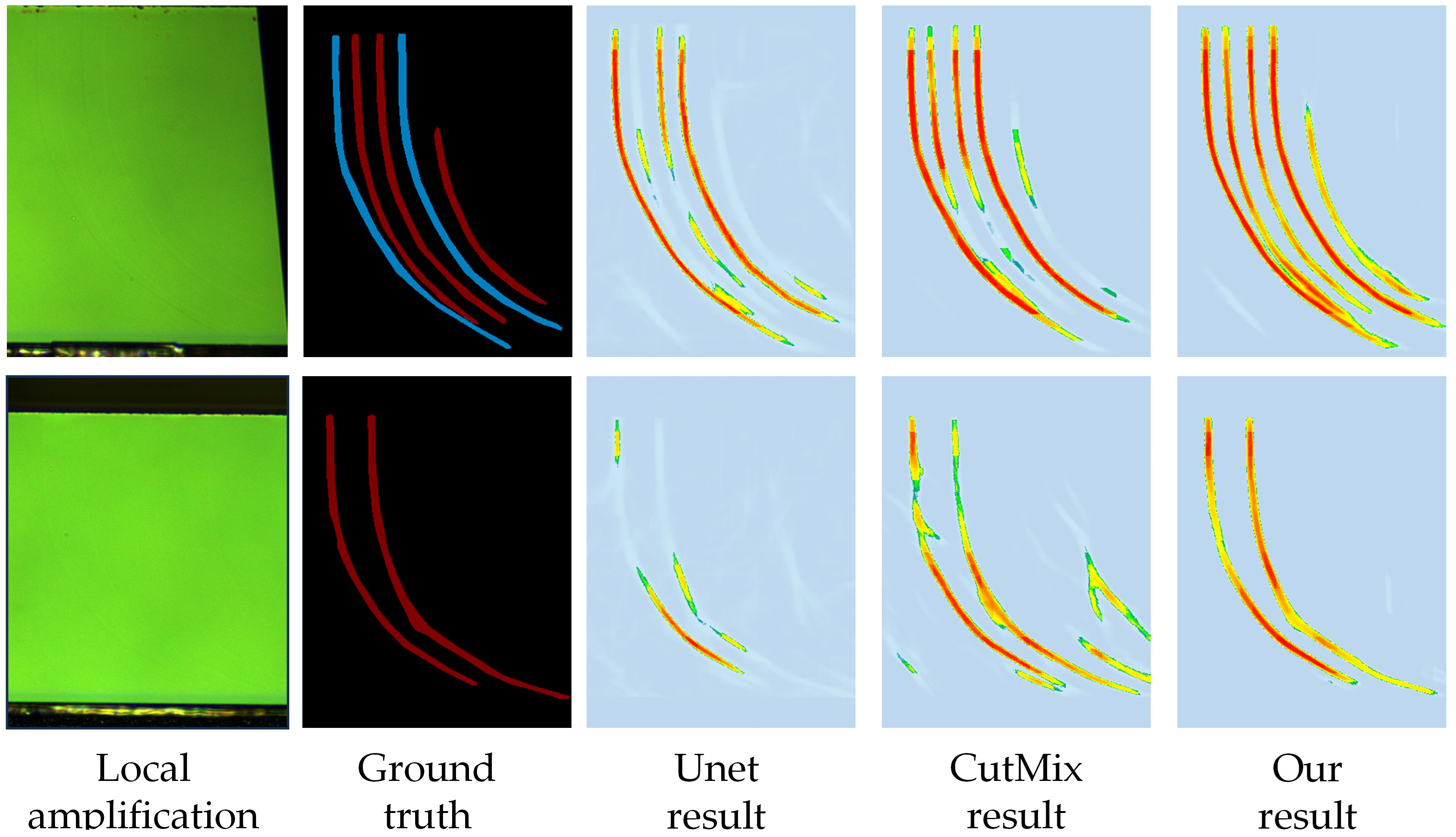}
\caption{Comparison of segmentation effects of shallow scratches}
\label{fig_6}
\end{figure}

\textbf{Shallow Scratch Detection Capability.} Ensuring consistent non-destructive detection of laser chip scratches primarily involves preventing the missed detection of shallow scratches. In actual detection, due to the high similarity between the representation of shallow chip scratches and the luminescent surface, and the significant lack of labeled datasets, traditional detection methods face two main challenges in shallow scratch detection. Firstly, incorporating readily available unlabeled data from the chip production process is essential for high-precision detection. However, the limited dataset support hampers traditional detection networks in identifying shallow scratches, making subsequent semi-supervised training impractical. Secondly, once basic segmentation of shallow scratches is achieved under supervised conditions, the inconspicuous nature of their features can lead to incorrect supervision signals in semi-supervised signal construction, causing the model to often misidentify them. To tackle the first challenge, TexRecNet improves shallow scratches detection by constructing a recursively optimized prediction process, which prioritizes foreground-background positional information during network prediction process. To avoid the generation of incorrect supervision signals, the method uses consistent texture entropy to evaluate the detection ability of the network for each chip image, focusing on the texture consistency in prediction sequences. This approach reduces the creation of supervision signals for hard to detect images, thereby minimizing erroneous signal generation. As shown in \cref{fig_6}, TexRecNet, using these methods, reduces both missed and false detections, thereby satisfying the practical demands of industrial applications.

\begin{figure}[!t]
\centering
\includegraphics[width=1.0\columnwidth,height=2.02in]{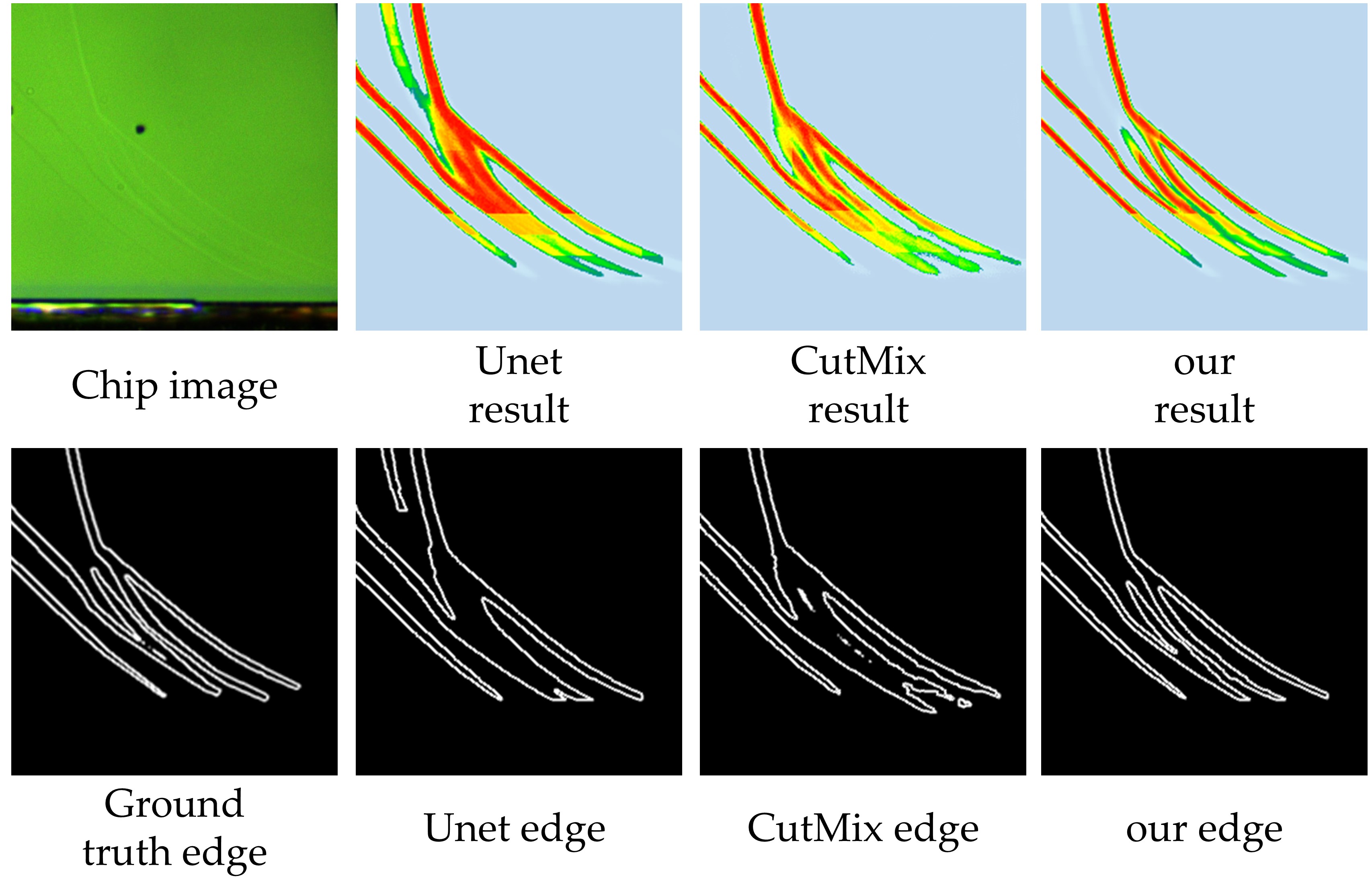}
\caption{Comparison of segmentation effects of deep scratches}
\label{fig_7}
\end{figure}

\textbf{Deep Scratch Detection Capability:} To validate the capability of the proposed method in detecting the edge structure of deep scratches, comparative experiments were conducted as illustrated in \cref{fig_7}. For an equitable comparison, both CutMix and the proposed method were trained using a $1/4$ data partition. It is observed that the proposed method significantly outperforms both the fully supervised Unet and the semi-supervised CutMix in capturing local texture details. This superiority is attributed to the proposed semi-supervised training strategy. Traditional semi-supervised training methods, which rely on image enhancement techniques, disrupt original image distribution to construct supervision signals. However, this approach is unsuitable for chip scratch detection, where the low signal-to-noise ratio exacerbates the disruption, leading to blurred local texture details. In contrast, the proposed consistency-constrained semi-supervised training method utilizes resampling and recursive optimization to generate multiple predictions for the same instance. By constructing consistency constraints, this method achieves lossless data augmentation and efficient semi-supervised training, thereby enhancing the detection capability of scratch positions and structures.

\subsection{Ablation}

\begin{table}
\centering
\renewcommand{\arraystretch}{1.4}
\caption{Comparison of IoU under different network inputs.}
\label{tab2}
    \begin{tabular*}{\linewidth}{@{\extracolsep{\fill}}ccccc}
    \toprule
    \toprule
    Model & supervised & 1082:2240(1:2) & 664:2658(1:4) & 369:2953(1:8) \\
    \midrule
    No ${{\widehat{x}}_{0}}$ & 69.3  & 73.2  & 71.5  & 68.4 \\
    ${{\widehat{x}}_{0}}=0$ & 68.5  & 72.0    & 70.4  & 68.1 \\
    Ours  & \textbf{71.7} & \textbf{75.6} & \textbf{74.5} & \textbf{72.4} \\
    \midrule
    Model & supervised & 1082:2240(1:2) & 664:2658(1:4) & 369:2953(1:8) \\
    \bottomrule
    \bottomrule
    \end{tabular*}%
\end{table}

\textbf{Effectiveness of the Recursive Optimization Structure:} To showcase the pivotal role of a recursive iterative network prediction process in enhancing the performance of the proposed model, specially the extraction of foreground-background positional information from previous recursions for use in subsequent ones, a series of experiments were designed. The experiments compared different input construction strategies under the same supervised data conditions: one strategy eliminated the position prior attention structure, providing no prior masks ${{\widehat{x}}_{0}}$, and the other used fixedly constructed prior masks ${{\widehat{x}}_{0}}=0$.The results, detailed in \cref{tab2}, indicate that the proposed method attained the highest foreground IoU and shallow scratch recall in four dataset partitions (supervised only, $1/2$, $1/4$, $1/8$). These results demonstrate that performance improvements from creating effective cyclic of positional signal significantly surpass other options. This is due to the fact that without positional priors, the conditional encoder data of the network relies solely on the original image, leading to fixed encoding outcomes. This fixed encoding fails to effectively incorporate previous stage outcomes to support the current segmentation task, resulting in decreased performance.

\begin{figure}[!t]
\centering
\includegraphics[width=1.0\columnwidth]{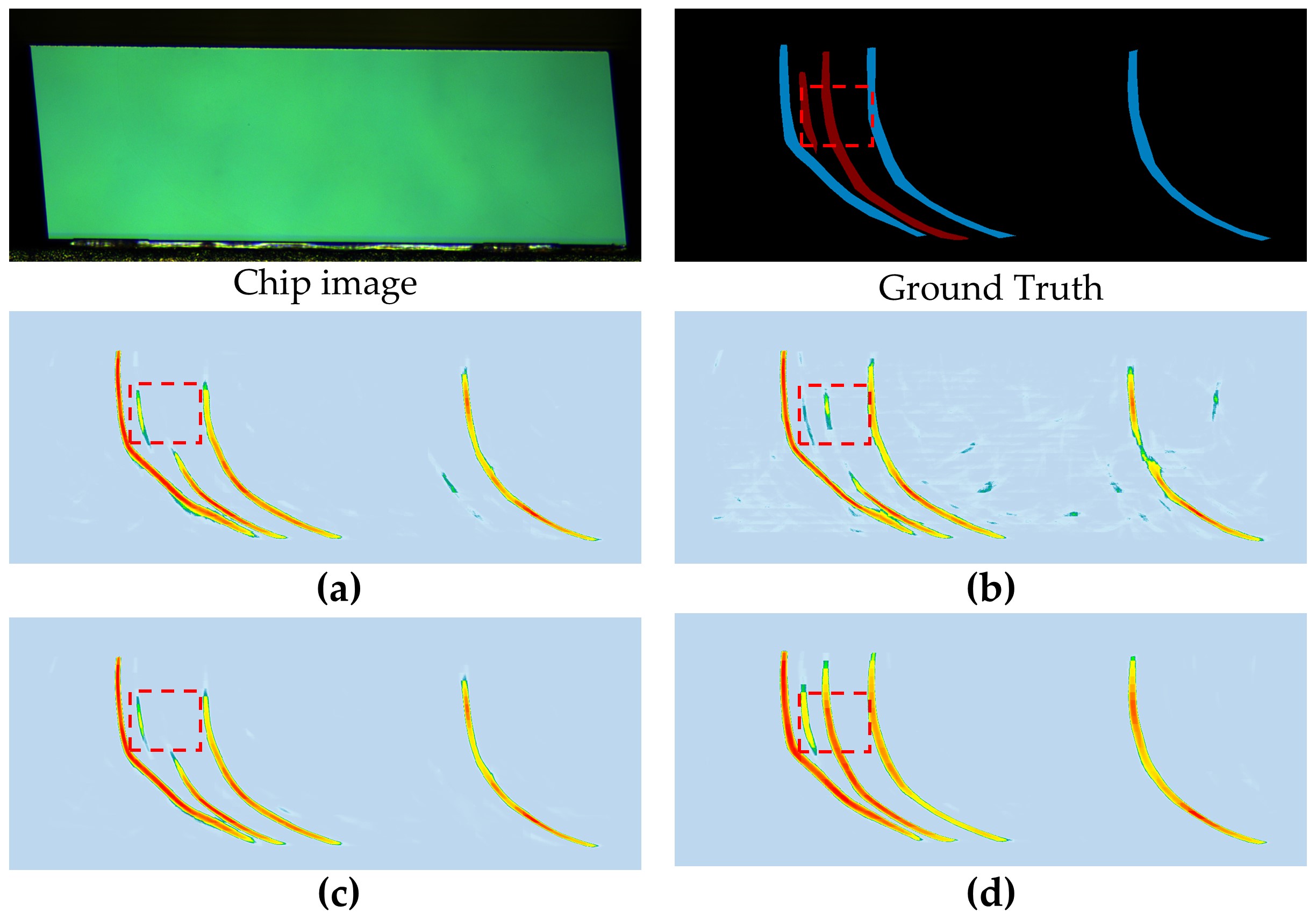}
\caption{Module Ablation (a) Removal texture entropy (b) Removal Resampling (c) Remove consistency constraints (d) Our method}
\label{fig_8}
\end{figure}

\textbf{Effectiveness of Semi-Supervised Training Components:} Further component ablation experiments were performed to precisely evaluate the effectiveness of each element in the TexRecNet semi-supervised method. These experiments were conducted using a $1/4$ data partition of the laser chip scratch dataset, consistently employing the network model proposed in this study as the basis. The results, as indicated in \cref{tab3} and \cref{fig_8}, reveal substantial performance reductions when texture entropy is substituted with image entropy or when the constraint on consistent texture entropy features in the supervision signal is eliminated. Specifically, these changes resulted in a decrease of 3.3$\%$ and 3.7$\%$ in IoU , and a 1.4$\%$ and 3.1$\%$ reduction in shallow scratch recall rate, respectively. As depicted in \cref{fig_8}, using image entropy, a measure of image disorder, to optimize supervision signals leads to inaccuracies due to its insufficient edge structure representation, resulting in improper constraints on these signals. Consequently, this leads to significantly narrower and more disordered network predictions of edges. Moreover, eliminating the random initial sampling strategy, an alternative to image enhancement, led to a 1.2$\%$ reduction in network prediction accuracy and increased errors in predicting the luminescent surface background of chips. In a control ablation experiment, removing the unsupervised training loss with consistency constraints and adopting conventional self-training methods resulted in a 3.1$\%$ decrease in prediction accuracy, thereby validating the effectiveness of the proposed method.

\begin{figure}[!t]
\centering
\includegraphics[width=1.0\columnwidth,height=2.42in]{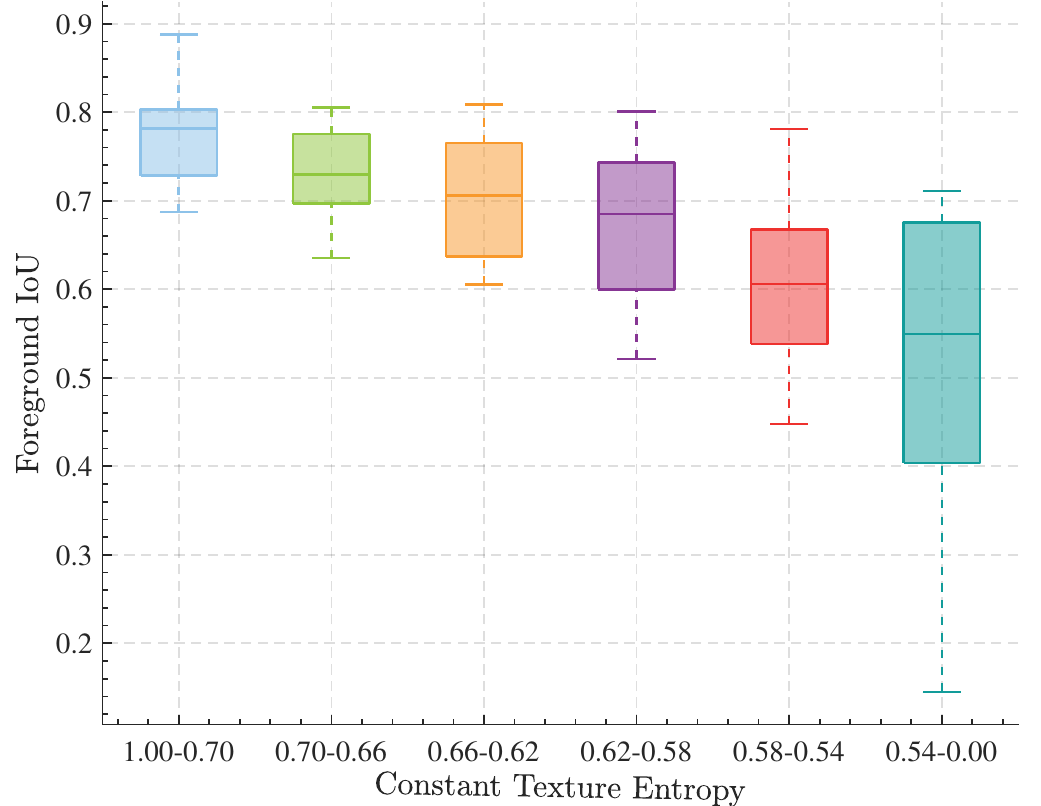}
\caption{Boxplot of IoU for consistent texture entropy features}
\label{fig_9}
\end{figure}

\begin{figure}[!t]
\centering
\includegraphics[width=1.0\columnwidth,height=2.42in]{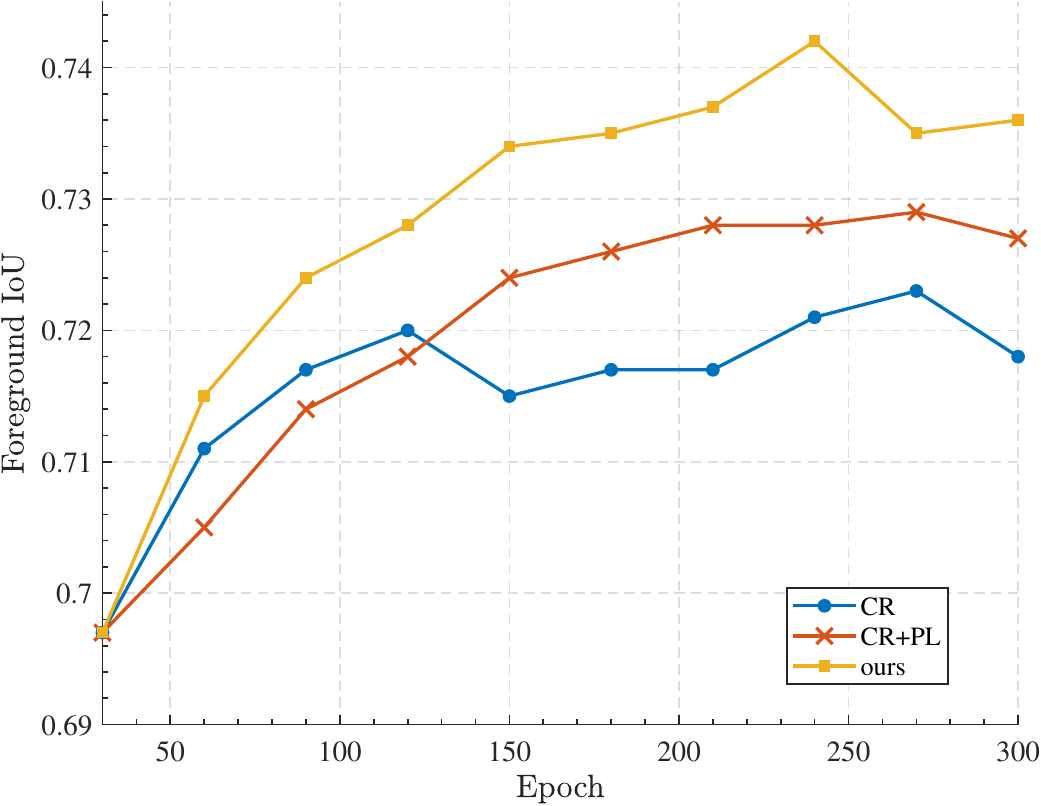}
\caption{Comparison of IoU Changes During the Training Process}
\label{fig_10}
\end{figure}

\textbf{Effectiveness of consistent texture entropy feature.} Based on the original structure texture distribution of chip scratches, the texture entropy index was proposed in this paper, and high-frequency shifts were extracted to derive consistent texture entropy features. To demonstrate the strong correlation between this representation and network cognitive confusion, statistical experiments were performed using the $1/4$ division training model described in this study. As shown in \cref{fig_9}, the consistent texture entropy features of the network prediction sequence were mainly distributed within the $[0.54,0.70]$ interval. Within this interval, there was a strong correlation between the representation and foreground IoU, a trend also observed in the $[0.70,1.00]$ interval. In the less sampled $[0.54,0.00]$ interval, despite a scattered distribution trend in the foreground IoU due to significant prediction inconsistencies, an overall correlation was still evident. Thus, it can be concluded that the consistent texture entropy features introduced  in this study effectively evaluate the prediction quality of the  network.

\begin{table}
\centering
\caption{Ablation research on each component of the semi supervised training strategy for TexRecNet.}
\label{tab3}
    \begin{tabular*}{\linewidth}{@{\extracolsep{\fill}}cccc|@{}c@{}c}
    \toprule
    \toprule
    \multicolumn{4}{c|}{TexRecNet on} & \multicolumn{2}{c}{664:2658(1:4)} \\
    \midrule
    ${{E}_{T}}$ & $\lambda $ & $N$ & ${{\mathcal{L}}_{u}}$ & IoU   & Shallow Recall \\
    \midrule
                   & \checkmark     & \checkmark     & \checkmark     & 71.2  & 68.8 \\
                   &                & \checkmark     & \checkmark     & 70.8  & 67.1 \\
    \checkmark     & \checkmark     &                & \checkmark     & 73.2  & 65.9 \\
    \checkmark     & \checkmark     & \checkmark     &                & 72.4  & 68.2 \\
    \checkmark     & \checkmark     & \checkmark     & \checkmark     & 74.5  & 70.2 \\
    \bottomrule
    \bottomrule
    \end{tabular*}
\end{table}

\textbf{Effectiveness of the Semi-Supervised Training Strategy.} To assess the effectiveness and efficiency of the semi-supervised training strategy presented in this paper, a comparative ablation with the consistency method was conducted. Experiments were conducted using a $1/4$ data division of the scratch dataset. The baseline model involved basic training with supervised data for 80 epochs. This was compared to unsupervised training strategies using Consistency Regularization (CR) and Consistency Regularization with Pseudo Labels (CR+PL). In the CR+PL method, Pseudo Label (PL) supervisory signals used the network's final segmentation to construct pseudo-labels. These signals were optimized through consistent texture entropy calculated from intermediate segmentations. As shown in \cref{fig_10}, the disruption of the original image distribution by CR in scratch images was significantly amplified by the low signal-to-noise ratio, caused repetitive training fluctuations and hindered accuracy improvement. The CR+PL method was modified to utilize information from the multi-stage prediction process, enhancing supervisory signals and improving overall accuracy, but at a slow rate of IoU increase. Our method, replacing traditional image enhancement methods like cropping and noise addition with random initial sampling, effectively improves the quality of supervision signals and calibrates the network optimization direction, achieving a markedly higher rate and upper limit of accuracy improvement than other methods. 
\section{Results}
This study presents a semi-supervised deep network based on consistent texture entropy recursive optimization, designed to overcome micro-scale, low contrast, and limited sample challenge in laser chip scratch detection. The recursively optimized network structure developed in the paper extracts foreground-background positional information from previous recursive processes, guiding subsequent recursive encoding. This iterative refinement of network predictions achieves high-precision detection of scratches under low contrast and micro-scale conditions. Additionally, a supervision signal construction method based on texture entropy is proposed. This method constructs supervision signals for unlabeled data using recursive prediction sequences and optimizes them by using differences in texture structures, thereby closely mirroring true signals, especially in edge structures. A semi-supervised training method based on iterative consistency constraints is also designed, achieving efficient semi-supervised training through recursive consistency loss and lossless data augmentation. Moreover, a dataset comprising challenging-to-detect scratch images on laser chip surfaces was constructed for validation experiments. The proposed method attained a 74.8$\%$ shallow scratch recall rate, marking a 33.6$\%$ improvement over Unet, thus demonstrating its superior performance in detecting challenging scratches on chips.

\bibliographystyle{IEEEtran}
\bibliography{IEEEabrv,reference}%

\vspace{11pt}

\vfill

\end{document}